%% file: main.tex
\documentclass{article}
\usepackage{ijcai23}

\usepackage{times}
\usepackage{soul}
\usepackage{url}
\usepackage[hidelinks]{hyperref}
\usepackage[utf8]{inputenc}
\usepackage[font=small, skip=0pt]{caption}
\usepackage{graphicx}
\usepackage{amsmath}
\usepackage{amsthm}
\usepackage{booktabs}
\usepackage{algorithm}
\usepackage{algorithmic}
\usepackage[switch]{lineno}

\include{pkgs}

\include{cmds}

\title{\paperTitle}

\author{
Abhijit Suprem$^1$
\and
Joao Eduardo Ferreira$^2$\And
Calton Pu$^1$
\affiliations
$^1$Georgia Institute of Technology, Atlanta, USA\\
$^2$University of Sao Paulo, Sao Paulo, Brazil
\emails
\{asuprem\}@gatech.edu
}

\begin{document}

\maketitle
\vspace*{-2em}
\input{abstract}

\input{intro}
\input{motivation}
\input{related}

\input{ufit}

\input{results}
\input{conclusions}

\bibliographystyle{named}
\bibliography{ijcai23}

\end{document}

%% file: pkgs.tex
\usepackage{endnotes,microtype,xspace,graphicx,fancyvrb,multirow}
\usepackage{pifont}
\usepackage{amsmath,amsopn}
\usepackage{listings}
\usepackage{subcaption}
\usepackage{mathrsfs}
\usepackage{booktabs}
\usepackage{listings}
\usepackage[table,xcdraw]{xcolor}
\usepackage[normalem]{ulem}
\useunder{\uline}{\ul}{}
\usepackage[mathscr]{euscript}
\usepackage{titlesec}

\usepackage[capitalize,noabbrev,nameinlink]{cleveref}
\usepackage{dsfont}
\usepackage{etoolbox}

%% file: cmds.tex
\newcommand{\squishitemize}{
\begin{list}{$\bullet$}
	{ \setlength{\itemsep}{0pt}
		\setlength{\parsep}{3pt}
		\setlength{\topsep}{3pt}
		\setlength{\partopsep}{0pt}
		\setlength{\leftmargin}{1.95em}
		\setlength{\labelwidth}{1.5em}
		\setlength{\labelsep}{0.5em} } }

\newcounter{Lcount}
\newcommand{\squishlist}{
	\begin{list}{\arabic{Lcount}. }
		{ \usecounter{Lcount}
			\setlength{\itemsep}{0pt}
			\setlength{\parsep}{3pt}
			\setlength{\topsep}{3pt}
			\setlength{\partopsep}{0pt}
			\setlength{\leftmargin}{2em}
			\setlength{\labelwidth}{1.5em}
			\setlength{\labelsep}{0.5em} } }
	
\newcommand{\squishend}{\end{list}}

\usepackage{xstring}
\newcommand{\PP}[1]{
\vspace{2px}
\noindent{\bf \IfEndWith{#1}{.}{#1}{#1.}}
}

\def\equationautorefname~#1\null{Eq. ~(#1)\null}

\def\Snospace~{\S{}}

\newcommand{\ttt}{\texttt}

\newcommand{\zerodisplayskips}{%
  \setlength{\abovedisplayskip}{0pt}%
  \setlength{\belowdisplayskip}{0pt}%
  \setlength{\abovedisplayshortskip}{0pt}%
  \setlength{\belowdisplayshortskip}{0pt}}
\appto{\normalsize}{\zerodisplayskips}
\appto{\small}{\zerodisplayskips}
\appto{\footnotesize}{\zerodisplayskips}

\newcommand{\ie}{\textit{i}.\textit{e}.\xspace}
\newcommand{\eg}{\textit{e}.\textit{g}.\xspace}

\renewenvironment{equation}{\vspace{-6pt}\begin{oldequation}}{\end{oldequation}\vspace{-6pt}}

\newcommand{\sys}{\textsc{UFIT}\xspace}

\newcommand{\red}[1]{\textcolor{red}{#1}}

\newcommand{\paperTitle}{Continuously Reliable Detection of New-Normal Misinformation: Semantic Masking and Contrastive Smoothing in High-Density Latent Regions}

\titlespacing\subsection{0pt}{5pt plus 4pt minus 2pt}{0pt plus 2pt minus 2pt}

%% file: abstract.tex
\begin{abstract}
Toxic misinformation campaigns have caused significant societal harm, \eg, 
affecting elections and COVID-19 information awareness. 
Unfortunately, despite successes of (gold standard) retrospective studies of 
misinformation that confirmed their harmful effects after the fact, 
they arrive too late for timely intervention and reduction of such harm. 
By design, misinformation evades retrospective classifiers 
by exploiting two properties we call new-normal: (1) never-seen-before novelty 
that cause inescapable generalization challenges for previous classifiers, 
and (2) massive but short campaigns that end before they 
can be manually annotated for new classifier training. 
To tackle these challenges, we propose \sys, which 
combines two techniques: semantic masking of strong signal keywords 
to reduce overfitting, and intra-proxy smoothness regularization 
of high-density regions in the latent space to improve reliability 
and maintain accuracy. 
Evaluation of \sys on public new-normal misinformation data shows 
over 30\% improvement over existing approaches on future (and unseen) 
campaigns. 
To the best of our knowledge, \sys is the first successful effort to 
achieve such high level of generalization on new-normal misinformation 
data with minimal concession (1 to 5\%) 
of accuracy compared to oracles trained with full knowledge 
of all campaigns. 
\end{abstract}

%% file: intro.tex
\section{Introduction}
\label{sec:intro}

Misinformation about the COVID-19 pandemic, dubbed infodemic \cite{infodemic} 
by WHO, has been recognized as causing harm to 
individuals and public health, 
as well as an erosion of confidence in institutions and technology. 
Concretely, fake news in social media has been found to influence 
elections \cite{alcott} and destruction of 5G cell phone towers \cite{fiveg}. 
To detect misinformation, state-of-the-art ML techniques such as fine-tuned 
LLMs \cite{misinfo}, weak-supervision \cite{mdaws}, mixture-of-experts 
\cite{moemisinfo} and others shown excellent accuracy, 
typically in retrospective studies using (gold standard) 
k-fold cross-validation on fixed annotated data sets.

\begin{figure}[t]
    \centering
    \begin{subfigure}{0.49\columnwidth}
        \includegraphics[width=\textwidth]{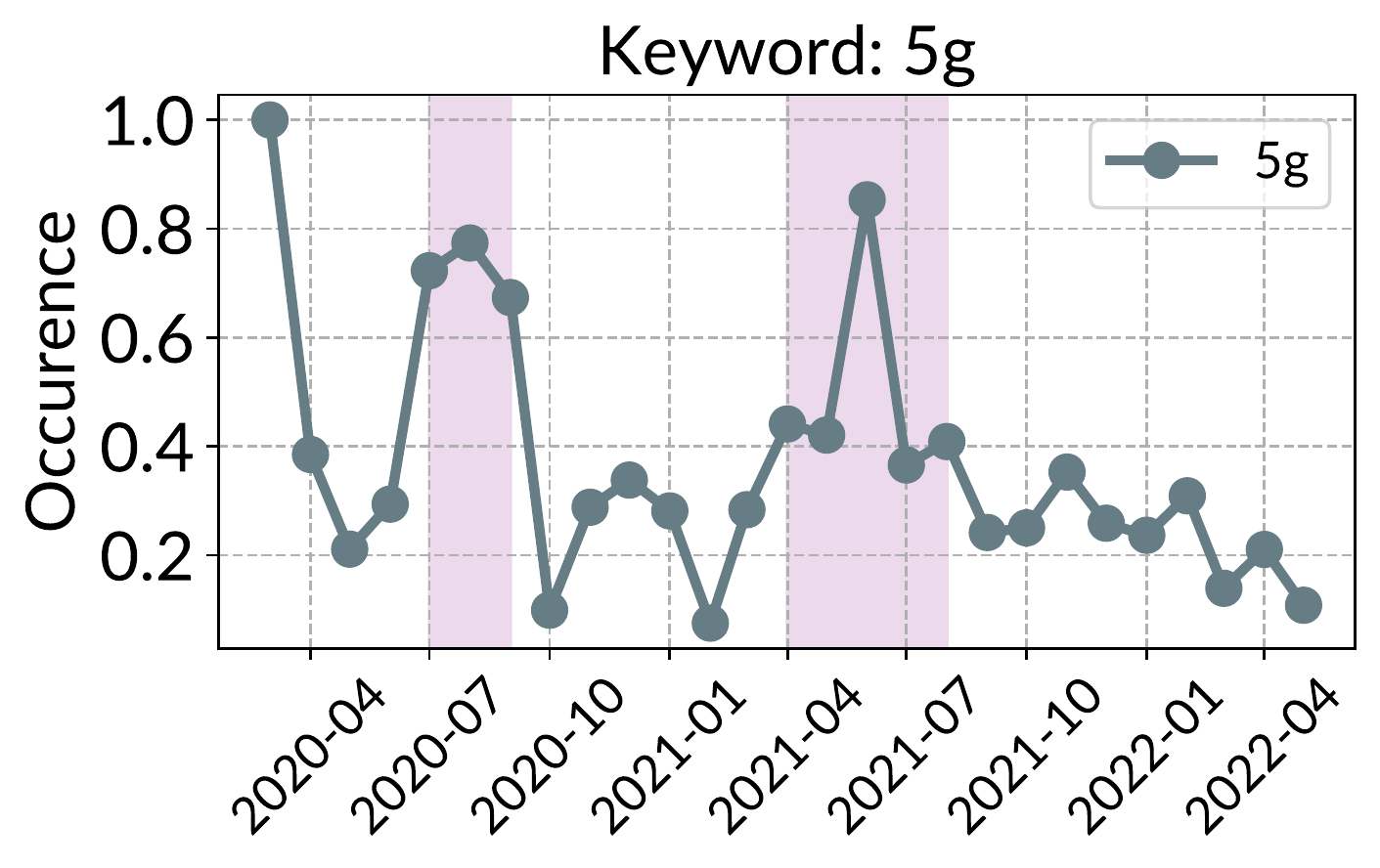}
        \label{fig:fiveg}
    \end{subfigure}
    \hfill
    \begin{subfigure}{0.49\columnwidth}
        \includegraphics[width=\textwidth]{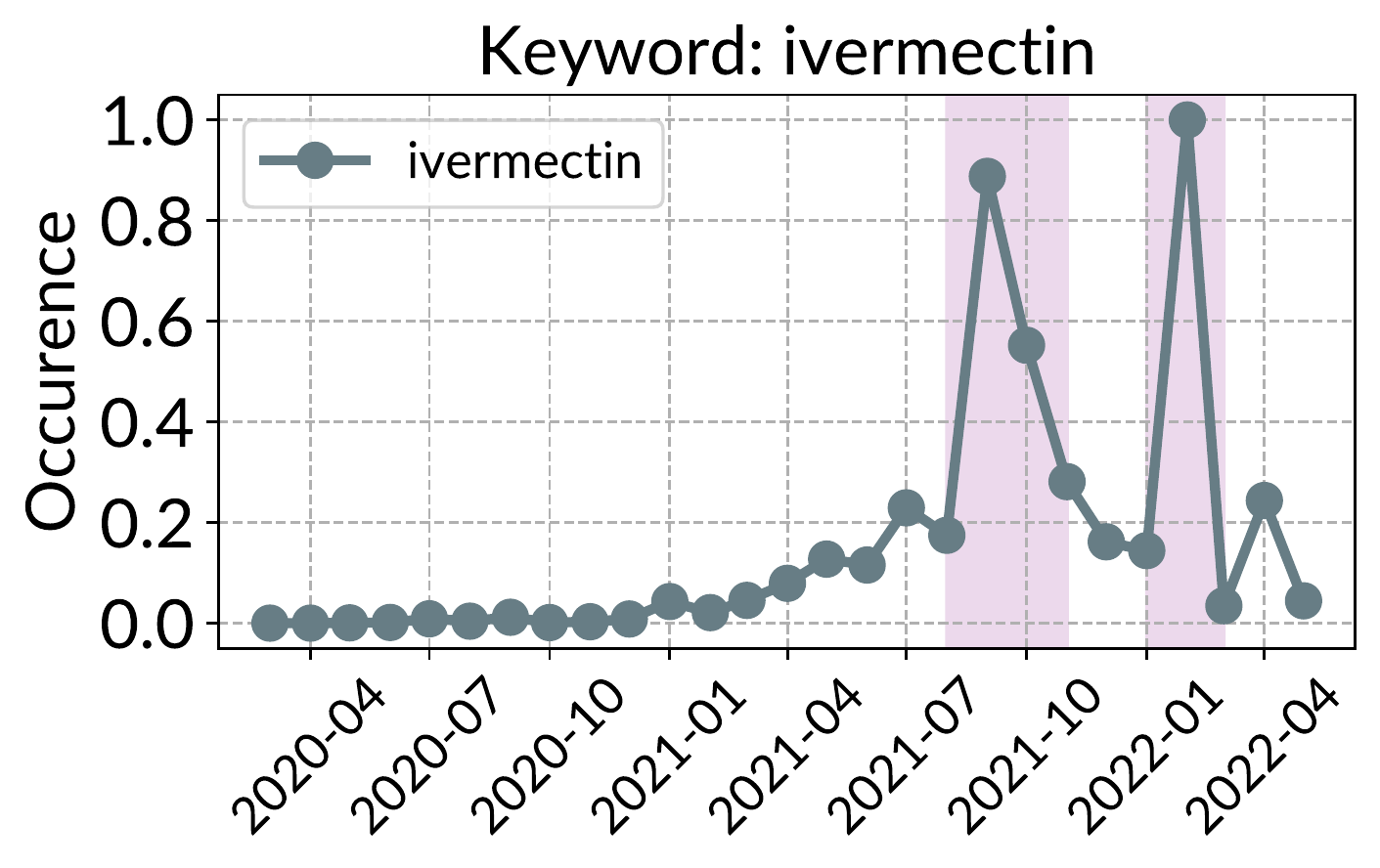}
        \label{fig:ivermectin}
    \end{subfigure}
    \vspace*{-1em}
    \caption{Occurrence of \ttt{5g} and \ttt{ivermectin} misinformation over time from \protect\cite{fnc}, a long-term dataset of COVID-19 misinformation on Twitter}
    \label{fig:occurence}
\end{figure}

Unfortunately, misinformation is designed to 
evade retrospective classifiers by exploiting two properties that we 
call \textit{new-normal}: 
(1) never-seen-before novelty, and 
(2) massive but short campaigns. 
On the first property, previously trained classifiers 
(without knowledge of never-seen-before novelty) have inevitable 
generalizability difficulties, shown in extensive evaluation 
of fixed dataset misinformation classifiers \cite{generalizability}. 
On the second property, fake news producers aim for social 
influence through massive injections;
however, the lifespan of misinformation campaigns 
is limited in the real world by the good work of fact checkers 
and authoritative sources (\eg, WHO, CDC on pandemics). 
\autoref{fig:occurence} illustrates the lifecycle of fake news 
campaigns (characterized by their theme), in short, massive waves 
that last a few weeks (using annotated data from \cite{fnc}). 
Given the inherent delay in (gold standard) manual annotations, 
we will need alternative approaches that can reliably detect 
new-normal misinformation early in their campaigns.

The above new-normal properties of misinformation introduce two significant 
and new research challenges previously missing in retrospective studies 
on fixed data sets. 
First, the continuous arrival of new-normal misinformation require  
capture of generalizable features to achieve sustained, 
reliable, and accurate detection of new-normal fake news across campaigns and novelty. 
This paper focuses on the feasibility of capturing generalizable latent 
features that are useful across multiple instances of never-seen-before 
novelty and massive short campaigns of misinformation through UFIT (\autoref{sec:sys}). 
The second challenge, to automate the incorporation of new knowledge in a 
timely manner to catch up with the new waves of new-normal misinformation 
in the real world, is the subject of ongoing research and future work. 
Since many readers are familiar with successful retrospective classification 
of fake news in fixed data sets, we start from an illustrative 
demonstration of the brittleness of typical fixed models 
when tested with new-normal fake news data (\autoref{sec:motivation}). 
More extensive studies have confirmed the brittleness of 
fixed classifiers \cite{generalizability,general2} of COVID-19 fake news datasets; 
\cite{nakedking} explores brittleness as a function of extensive fine-tuning on sentiment analysis.

\PP{Contribution 1: UFIT}
We propose \sys to exploit high-density regions in the 
latent space, which occur naturally in clustered data such as 
new-normal fake news campaigns. 
Due to their high signal-to-noise ratio, overfitting often arises 
in these high-density clusters. 
\sys integrates 2 techniques to balance overfitting with 
underfitting through judicious control of model components 
in these high-density regions: (a) \textbf{semantic masking} of strong-signal 
keywords that correspond to the high-density clusters in training 
data latent space, and (b) \textbf{intra-proxy embedding regularization} 
within the high-density regions of the latent space 
to improve local smoothness and reduce overfitting.
We cover related works for improving generalizability, which we call 
controlled underfitting, in \autoref{sec:cufit}.

\PP{Contribution 2: Experiments}
We evaluate \sys{'s} reliability and accuracy with extensive experiments 
on several groups of semantically similar new-normal datasets 
that exhibit both never-seen-before novelty and massive short campaigns. 
A key innovation is the experimental setup, since we ensure existence 
of new-normal properties by integrating ordered datasets that 
contain natural and realistic distribution shifts for misinformation. 
We show that \sys significantly improves reliability on such new-normal distributions. 
On the curated EFND collection of 11 COVID-19 misinformation datasets from \cite{generalizability}, 
we can improve the new-normal distribution accuracy by almost 25-30\% 
relative to classifiers without controlled underfitting. 
On the yearly dataset releases of NELA \cite{nela}, a general misinformation collection from 2018-2022, 
\sys classifiers improve on non-UFIT classifiers by $\sim$30\% across multiple years.

%% file: motivation.tex
\section{Fixed Model Brittleness to Novel Data}
\label{sec:motivation}

\begin{figure}[t]
    \centering
    \begin{subfigure}{0.49\columnwidth}
        \includegraphics[width=\textwidth]{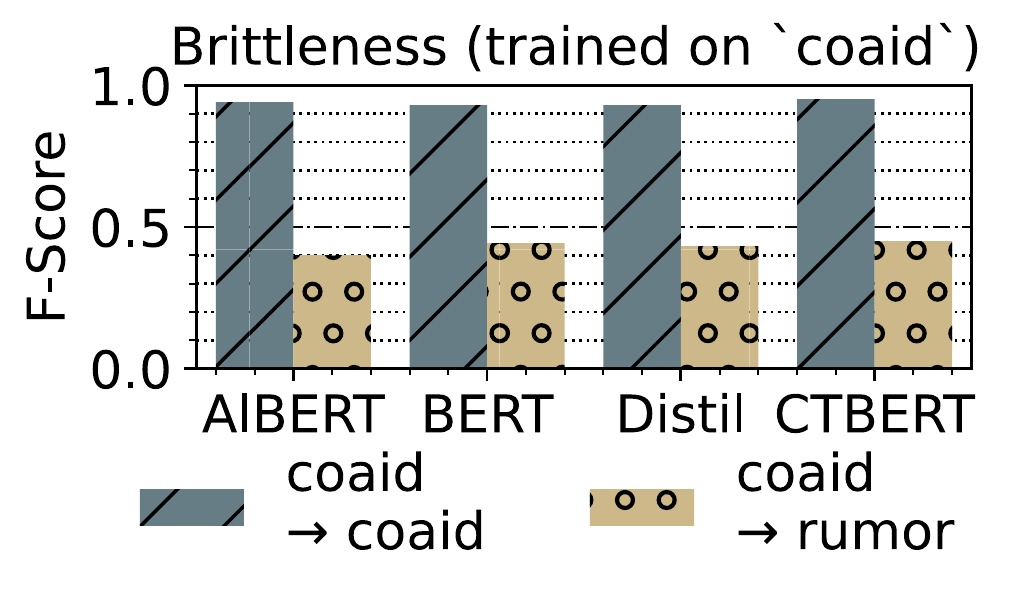}
        \label{fig:brittleCoaid}
    \end{subfigure}
    \hfill
    \begin{subfigure}{0.49\columnwidth}
        \includegraphics[width=\textwidth]{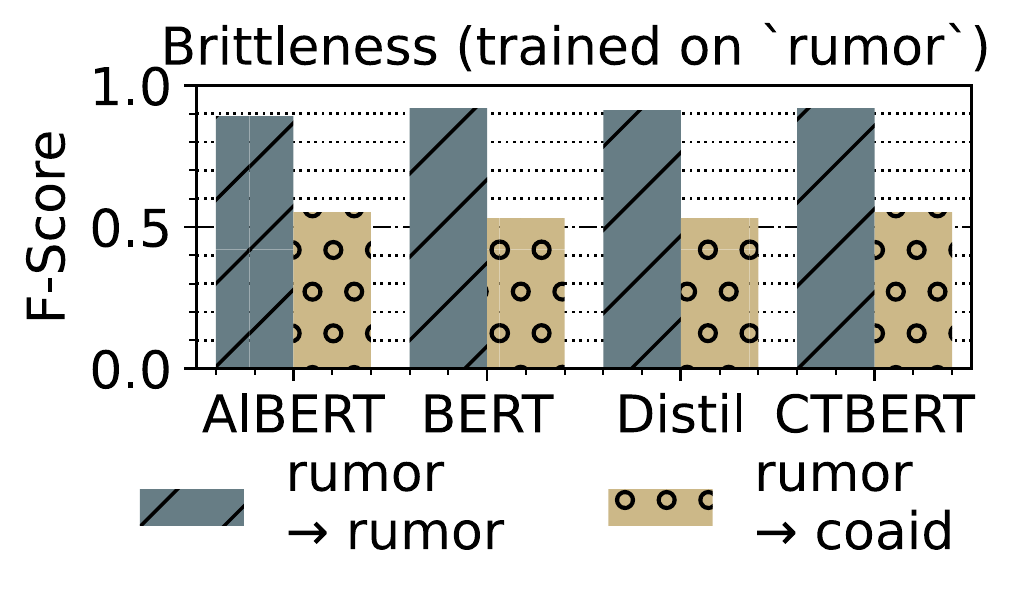}
        \label{fig:brittleRumor}
    \end{subfigure}
    \vspace*{-1em}
    \caption{\textbf{Brittleness experiment}: We train on \ttt{rumor}
     and test on \ttt{rumor} (ID) and on 
     \ttt{coaid} (NNAD), and vice versa. 
     Classifiers perform much worse on NNAD 
     even though the datasets are similar.}
    \label{fig:brittle}
\end{figure}

When tested with in-distribution (ID) data within fixed data sets, 
fixed models achieve excellent performance \cite{mdaws,misinfo,midas,wilds}. 
However, they exhibit brittleness when faced with out-of-distribution data (OOD) \cite{wilds,generalizability,templama}. 
To remedy this problem, some fixed models incorporate stationary sub-models 
(\eg well-defined statistical distributions) of originally OOD data, 
effectively augmenting the fixed model when the statistical assumptions of 
OOD hold true. 
Unfortunately, the never-seen-before novelty of the new-normal is 
generated by fake news producers to avoid such predictable patterns \cite{infodemic}
in the mentioned massive short campaigns (\autoref{fig:occurence}). 
We refer to this distribution shift with new-normal data as new-normal arbitrary distribution (NNAD). 
We note that `arbitrary' includes both random and non-random distributions, 
since NNAD is under adversarial control of fake news producers.

Consider 2 COVID-19 misinformation datasets from EFND \cite{generalizability}: 
\ttt{rumor} and \ttt{coaid}, containing headlines (through twitter) 
and tweets with COVID-19 misinformation and factual information collected in 2020. 
This is an example of NNAD, since \ttt{rumor} and \ttt{coaid} are never-seen-before novelty w.r.t. each other.

\begin{figure}[t]
    \centering
    \begin{subfigure}{.85\columnwidth}
        \includegraphics[width=\textwidth]{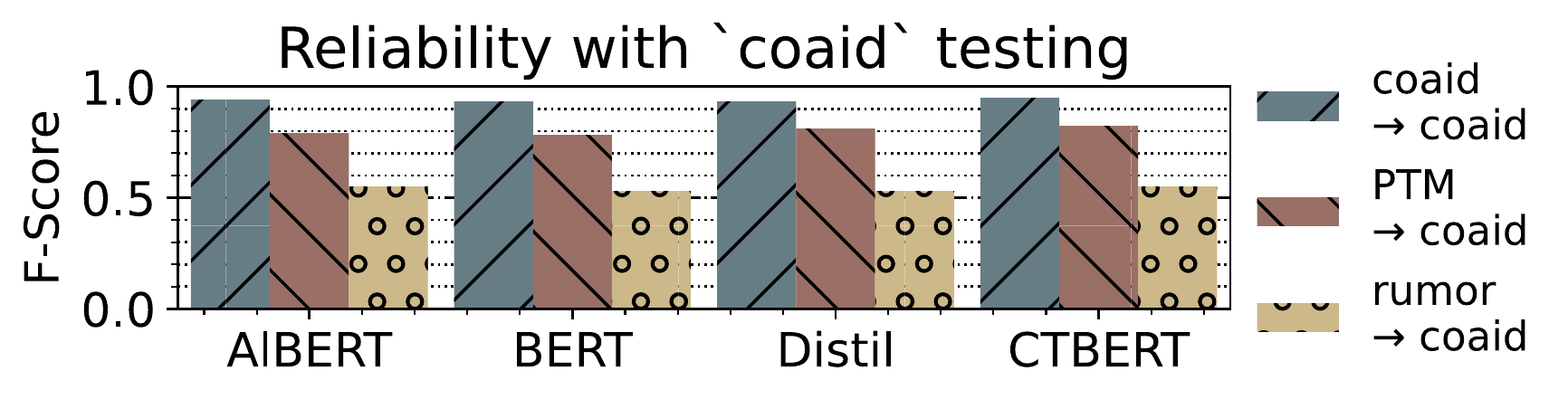}
        \label{fig:robustCoaid}
    \end{subfigure}
    \\[-1em]
    \begin{subfigure}{.85\columnwidth}
        \includegraphics[width=\textwidth]{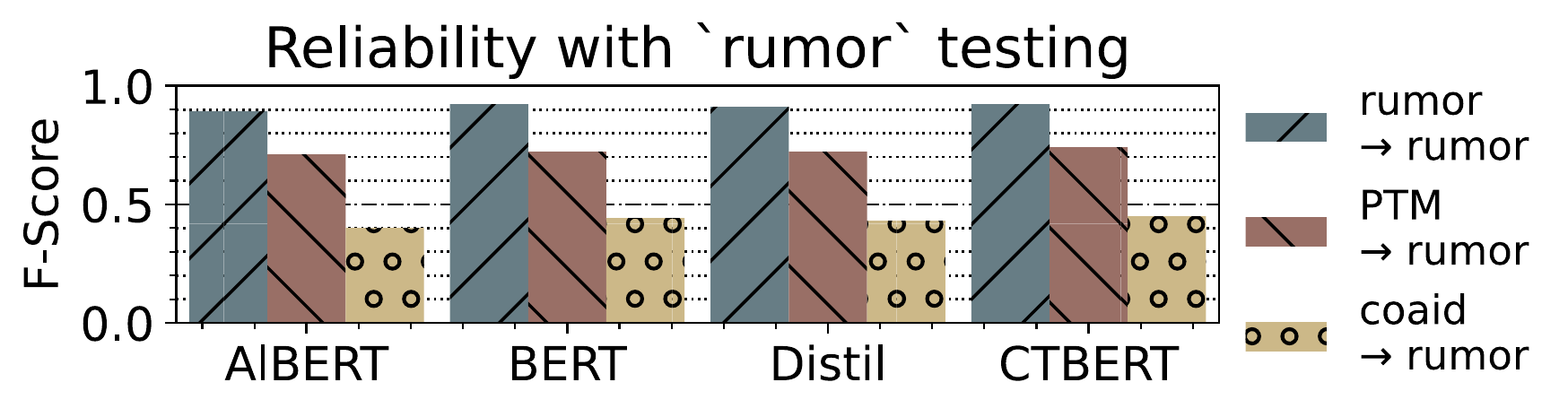}
        \label{fig:robustRumor}
    \end{subfigure}
    \vspace*{-1em}
    \caption{\textbf{Robustness experiment}: The pre-trained model (PTM) 
    without any fine-tuning perform reasonably well. 
    Once classifiers are fine-tuned, their performance 
    on NNAD data is worse than the PTM.}
    \label{fig:robust}
\end{figure}

\PP{Brittleness} \autoref{fig:brittle} shows NNAD evaluation: fixed classifiers trained on 
\ttt{rumor} and tested on \ttt{coaid}, plus vice versa, show accuracy 
decline of $\sim$20-30\% across all tested backbones. 

\PP{Loss of Reliability} \autoref{fig:robust} includes some well-known pre-trained models (PTM) without fine-tuning. 
Compared with the previous fixed classifiers, the PTMs performed somewhat better, at about 20\% accuracy loss. 
The brittleness of fixed classifiers illustrate the widespread generalizability 
difficulties of fixed models from \cite{generalizability,templama}. 
A more detailed discussion with further evidence is included in supplementary work.

%% file: related.tex
\section{Related Work}
\label{sec:related}

\subsection{New-Normal Datasets}
\label{sec:newnormal}

Several datasets have recently been proposed to study the 
new-normal nature of real-world distributions and knowledge. 
NELA\cite{nela} releases yearly snapshots of misinformation on 
news and twitter collected from multiple whitelisted and blacklisted accounts. 
TempLama \cite{templama} is a recent temporal fact database to 
study LLMs' ability to distinguish `permanent' (number of continents) and 
`temporally-sensitive' (President of US) facts. 
FNC\cite{fnc} provides a 24-month annotated multilingual dataset of COVID-19 
misinformation on twitter through multiple misinformation campaigns. 
WILDS\cite{wilds} is a benchmark for distribution shift testing 
with 10 real-world dataset collections. 

A fundamental difference arises between fixed, synthetic data sets 
and NNAD: fixed data can be captured by 
stationary models due to their finite size, but NNAD contain new-normal that evade 
stationary models. 
Therefore, evaluation experiments on synthetic novelty, \eg, as provided by 
some subsets of WILDS, would have generalizability difficulties \cite{temporalptm}. 
From this perspective, NNAD can be considered a general case of 
novel class detection (\eg, in data streams with concept drift) \cite{khandrift}.

\subsection{Distribution Shift and Concept Drift}
\label{sec:distshift}
Distribution shift (and the distinctions between ID, OOD, and NNAD) 
has roots in concept drift\cite{gama}, which identifies 2 variants: 
real concept drift (never-seen-before and thus unpredictable novelty, or changes in posterior probabilities) 
and virtual concept drift (changes in distribution priors without affecting posterior). 
Taking cues from recent works \cite{wilds}, we use distribution shifts, real concept drift, 
and NNAD interchangeably. 
Distribution shifts are a significant problem for sustainable and reliable 
machine learning \cite{templama,wilds,temporalptm}. 
However, as noted in \cite{wilds}, most existing work on distribution 
shift focuses on well-defined isolated test-beds for 
domain generalization and do not represent realistic distribution 
shifts where subpopulations can shift causing significant 
classifier degradation with semantically similar distributions (this is the NNAD case in \autoref{sec:motivation}). 
The same is true for misinformation, where there is recent emerging work 
on the NNAD case of multiple novel and ephemeral variants 
of misinformation campaigns: \cite{generalizability} surveys 11 
COVID-19 datasets to measure classifier brittleness, 
\cite{general2} evaluates transformer backbones on misinformation datasets, 
and \cite{mdaws} proposes a embedding invariance method 
for identify common contextual signals across fake news domains.

\subsection{Controlled Underfitting}
\label{sec:cufit}
We refer to \textit{controlled underfitting} as any technique 
employed to mitigate overfitting in neural networks; 
as such, controlled underfitting has a rich history. 
The transformer architecture\cite{transformer} and several variants of 
pretrained bases are ubiquitous in modern NLP, 
and have continuously employed randomized masking to 
reduce overfitting\cite{rwm,robust,nakedking}. 
The goal for masking, to learn contextual embeddings and associations 
instead of memorizing the input, clearly fits the controlled underfitting 
paradigm; the `control' comes from the masking probability.  
Similarly, token replacement/deletion and next sentence prediction \cite{bert} also reduce 
overfitting in pre-trained LLMs. 
To our knowledge, Masker\cite{masker} is the only data augmentation 
underfitting technique that proposes masking strong signals 
during the fine-tuning process as well as the pre-training process to reduce brittleness.

In addition to data augmentation, several training methods 
and regularization are also effective at controlled underfitting: 
training tricks and heuristics such as warmup learning rate \cite{tricks,regularization}, 
Mixout regularization as a replacement for dropout \cite{mixout}, 
FreeLB as an adversarial training technique \cite{freelb}, 
and SMART as a parameter update optimizer \cite{smart} are all techniques to 
improve stability of training and ID test accuracy.

%% file: ufit.tex
\section{\protect{\sys}}
\label{sec:sys}


\begin{figure}[t]
    \centering
    \begin{subfigure}{0.49\columnwidth}
        \includegraphics[width=\textwidth]{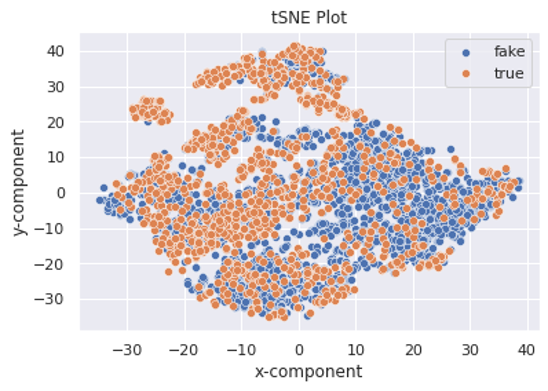}
        \caption{Pre-trained Model}
        \label{fig:ptm}
    \end{subfigure}
    \hfill
    \begin{subfigure}{0.49\columnwidth}
        \includegraphics[width=\textwidth]{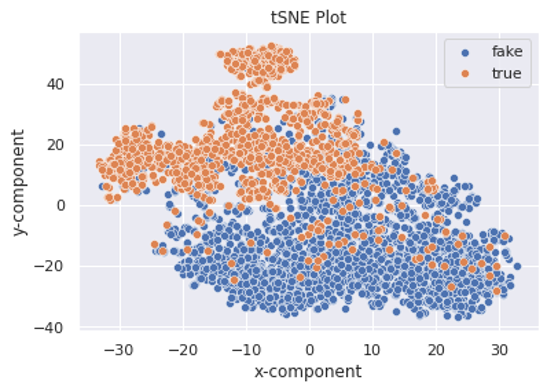}
        \caption{Fine-tuned Model (FT)}
        \label{fig:ft}
    \end{subfigure}
    \\
    \begin{subfigure}{0.49\columnwidth}
        \includegraphics[width=\textwidth]{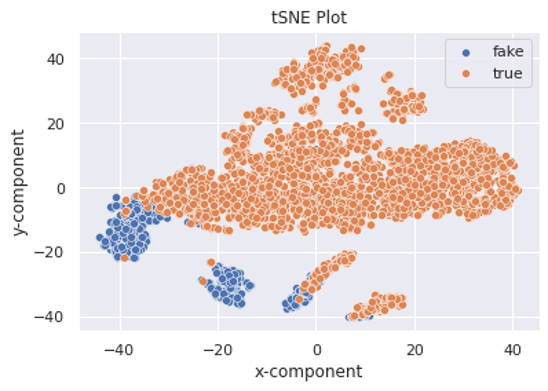}
        \caption{Semantic Masking}
        \label{fig:sm}
    \end{subfigure}
    \hfill
    \begin{subfigure}{0.49\columnwidth}
        \includegraphics[width=\textwidth]{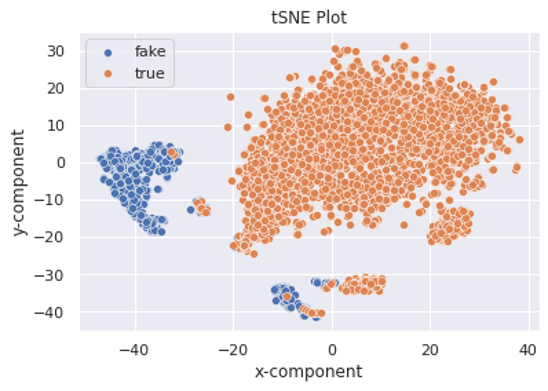}
        \caption{Masking+Smoothness (UFIT)}
        \label{fig:smreg}
    \end{subfigure}
    \caption{The latent space (tSNE of EFND embeddings) in an NNAD setting shows impact of masking and regularization: Masking 
         identifies important proxy regions and smoothness regularization 
          generates more separable clusters.}
    \label{fig:embeddings}
\end{figure}

\subsection{Design Rationale}
\label{sec:rationale}

When applying controlled underfitting to the new-normal data sets,
the two properties of never-seen-before novelty and massive, short campaigns 
suggest differentiated consideration of carefully separated dense regions 
of the latent space (discussed with respect to our dataset choices in \autoref{sec:datasets}). 
\sys exploits existence of high-density disjoint clusters in the latent space \cite{odin,trust}
to learn more separable projections of the input that are reliable for classification across
new-normal properties. 
This separation of clusters is illustrated in \autoref{fig:embeddings}, where \sys (\autoref{fig:smreg}) achieves the best separation of 
 embeddings for \ttt{coaid} in the NNAD setting of the Motivating Experiments in \autoref{sec:motivation}.

We illustrated \sys{'s} goal of improved reliability from controlled underfitting in an NNAD setting (\autoref{fig:robust}), 
\sys achieved higher accuracy than PTMs and fine-tuned classifiers. 
In ID experiments (\autoref{fig:brittle}), fine-tuned classifiers achieve high accuracy 
by overfitting on strong signals in ID training data. 
In the NNAD setting, previous (ID) strong signals are replaced (often adversarially) 
by never-seen-before (novel) strong signals. 
This replacement causes the 40\% degradation in accuracy of overfit fine-tuned classifiers in \autoref{fig:brittle}.

To achieve \sys{'s} goal, the first technical component: 
semantic masking (\autoref{sec:masking}), attempts to capture the replacement of 
previous strong signals in NNAD by masking the strong signals themselves from the previous campaign. 
As shown by the PTM in \autoref{fig:robust} (which are trained with masked language modeling), 
masking in general can improve reliability; we augment this with targeted masking of strong signals. 
%
The second technical component of \sys, intra-proxy embedding regularization 
(\autoref{sec:regularization}), smooths the regions around the masked signals to 
improve the latent space projection. 
By integrating the two components, \sys smooths out the impact from the 
loss of previous strong signals to achieve reliable detection of future new-normal campaigns.

\subsection{High-density Sets}
\label{sec:proxies}
To identify high-density clusters, we look to topological level 
set estimation as applied to latent space clustering in \cite{trust} and \cite{odin}. 
They introduce the notion of a $\alpha$-density set of 
a class as the training set of the class after filtering out an $\alpha$ fraction of  
low-density points. 
To find the high-density set for each class, we let $\alpha=0.5$ 
and use KMeans to filter the $1-\alpha$ samples closest to the cluster 
centroid. 
However, this belies a more complex situation: the number of 
classes in training data are not representative of the 
number of disjoint clusters, also known as \textbf{proxies} in the latent space \cite{proxynca}; 
in fact, the latent space contains multiple disjoint proxies 
that are projected to a single class through non-linear dense layers \cite{tricks}.

This leads to our key insight: by estimating \textit{class agnostic} 
high-density sets in the latent space, we can apply proxy-specific underfitting to 
improve reliability and reduce over-fitting directly on the 
strong signals in high-density proxies, 
\textit{while ensuring capture of weak yet important signals in the low-density regions between proxies}.

We identify proxies by first fine-tuning a PTM (we use AlBERT for most experiments) 
with the standard classification head and obtain embeddings for the training data. 
Then, we perform KMeans clustering on each known class, sweeping through 
different values of K with $p$ multiples of $C$ classes (e.g. for 2 classes, we can cluster with $K={2,4,6,\cdots,2p}$) 
to find the optimal number of clusters (visually with ELBOW metric or automated with gap statistic method). 
This yields us $p\cdot C$ proxies, each with a subset of the training data.

Then, we apply proxy-specific controlled underfitting. 
First, we perform \textbf{Semantic Masking}, where for each proxy, we mask popular keywords on the 
corresponding samples to reduce over-reliance on them. 
This is an intuitive choice since we can augment the agnostic 
keyword choices with domain expertise, \eg the topic keywords of \texttt{5G} and \ttt{Ivermectin} 
in \autoref{fig:occurence}. 
Masking, however, can leave `holes' in the latent 
space within proxies where classifiers have not captured 
enough data to perform good projections. 
The second component then is \textbf{Intra-proxy Regularization} to complement masking 
by increasing smoothness to improve 
the clustering quality in the latent space and increase separability.

Our evaluation experiments in \autoref{sec:results} show the contributions of each component, 
and their combination that outperforms considerably either component, 
suggesting the potential value of cluster-specific controlled underfitting as a general approach. 

\subsection{Semantic Masking}
\label{sec:masking}
Intuitively, the keywords we mask should correspond 
to strong signals in the identified latent space proxies.

\PP{Extract Strong-Signal Keywords}
We use the fine-tuned classifier to generate the final attention 
layer for each training sample $x_i\in \mathscr{D}_{k}$, where $k$ 
is a single proxy cluster. 
Then, given vocabulary size $l$ in proxy $k$, for each word $w_l^k\in{x_i}_{i=1}^{|\mathscr{D}_{k}|}$, we compute
the word attention score $a_{kl}^w$ given sample $x_i$ with length $|x_i|$
and containing tokens $t_j$ with token attention $a_j^t$:

\begin{equation}
\label{eq:attscore}
a_{kl}^w = \sum_{x_i\in\mathscr{D}_{k}}\bigg( \tau \sum_{j=1}^{|x_i|} \mathds{I}(t_j\in w_k)a_j^t \bigg)
\end{equation}


Each sample's token attention scores ${a_j^t}_{j=1}^{|x_i|}$ are softmax normalized for
that sample. 
$\tau_{ki}$ is a sigmoid temperature scaling parameter that adjusts the normalized attention
scores by taking into account sample length $|x_i|$ as well as the median of
all sample lengths $\hat{\tau}_k$ in $\mathscr{D}_k$:

\begin{equation}
    \label{eq:temperature}
    \tau_{ki} = \bigg(  1 + \exp \frac{-|x_i|-\hat{\tau}_k}{\sqrt{\hat{\tau}_k}}\bigg)^{-1}
\end{equation}

This dampens keyword attention scores in short 
sentences and emphasizes them in 
long sentences with heavy-tailed attention distribution after softmax normalization. 

\PP{Aggregate and Group Keywords}
The prior step computes top-ranked words in each proxy cluster. 
We group all top-$k$ attentions of proxies in 
the same class to obtain aggregate class-specific keywords, 
and use the top-1 keyword from each proxy. 
%
%
By masking these keywords, we force the classifier to focus on surrounding text to learn contextual 
signals. 
Different from Masker\cite{masker}, we have added the $\tau$ to deal with variable 
sequence lengths, and we have used proxies to obtain a more local estimate of strong signal keywords.

\PP{Semantic Masked Language Modeling}
Next we re-train the PTM with semantic masked language modeling (mlm) 
loss to adjust the latent embeddings so they reflect the distribution of 
the training data without overfitting on strong signal keywords (analogous to random erase augmentation \cite{tricks}). 
For each sample, we mask keywords for that sample's class label, 
plus additional random words to hit the usual mlm probability of 0.15. 
The mlm loss is the standard cross-entropy loss $\mathscr{H}$ on the masked token 
prediction given sample $x_i$, the $j$-th token prediction $f_j$, and ground truth token $t_j$:

\begin{equation}
    \label{eq:lmlm}
\mathscr{L}_{MLM} = \mathscr{H}(f_j(x_i), t_j)
\end{equation}

\PP{Fine-tune classifier}
Next we fine-tune with a classification head, halving the learning rate for the embedding layers 
relative to the classification head to improve convergence and avoid aggressive updates \cite{smart}. 
We combine masked reconstruction loss $L_{MLM}$ on the decoder, 
task specific loss $L_{Task}$ on classification head (covered in \autoref{sec:regularization}), 
and mask regularization on the classification head $L_{KL}$. 
For the mask reconstruction loss, we use \autoref{eq:lmlm}; 
for mask regularization, we apply context masking to mask of all words \textit{except} keywords for some samples in a batch \cite{masker}. 
We minimize KL divergence loss between predictions and a uniform distribution, 
so that the classifier further underfits the strong signal keywords ($f$ is the classification head): 

\begin{equation}
\mathscr{L}_{KL} = D_{KL}(\mathscr{U}(C) || f(x_i))
\label{eq:lkl}
\end{equation}

Then, our loss for fine-tuning is (with weight $\lambda$):

\begin{equation}
\mathscr{L}_{FT} = \mathscr{L}_{Task} + \lambda_{MLM}\mathscr{L}_{MLM} + \lambda_{KL}\mathscr{L}_{KL}
\label{eq:lft}
\end{equation}

Different from Masker, we use proxies  in the latent space to extract local  
keywords corresponding to high-density sets. 
The benefit is that we extract strong-signal keywords only from 
regions of the training data the classifier is likely to overfit 
on (\ie the high-density regions), while ignoring the 
low-density regions with limited chance of overfitting.

\subsection{Intra-Proxy Regularization}
\label{sec:regularization}
After semantic masking, the quality of projections in the latent space is reduced, 
since we have masked important keywords. 
To address this, we combine 
\autoref{eq:lft} with a smoothing regularization loss, 
where our final fine-tuning loss takes 
the form: $L_{FT}=L_{Task}+L_{Smooth}$; $L_{Task}$ is the softtriple loss \cite{softtriple,softtriple2}.

We use softtriple, common in re-id, because of an intuitive insight 
that misinformation detection is similar to re-id. 
That is, both re-id task and misinformation campaigns contain 
high-velocity short-lived events (previous unseen vehicles/campaigns passing through a camera/stream). 
%
What we want is to cluster misinformation on features 
that are common across campaigns, while avoiding the strong-signal features on individual, 
short-lived misinformation. The latter is accomplished with Semantic Masking. 
We accomplish the former with softtriple, which combines triplet and smoothed softmax loss
to project embeddings into $k$ clusters (\eg proxies) per class $C$:

\begin{equation}
    \mathscr{L}_{Task} = -\log  \frac{\exp (\lambda (\mathcal{S}_{i, y_i} - \delta))}{ \exp \big(  \lambda (\mathcal{S}_{i, y_i} - \delta) \big)  \sum_j^{kC} \exp (\lambda\mathcal{S}_{i,j})   }  
    \label{eq:ltask}
\end{equation}

Here, $\lambda$ is a smoothing parameter, typically $\lambda=20$ in \cite{softtriple};
$\delta$ is a margin constraint for tie-breaking in proxy assignment, typically  
$\delta=0.01$, and $\mathcal{S}_{i,j}$ is the proxy assignment for sample $i$ to proxy $j$ of class $C$ ($y_i$ is the predicted class).
$\mathcal{S}$ computes similarity between $k-$th proxy and sample $x_i$; we defer a thorough explanation to \cite{softtriple}: 

\begin{equation}
    \mathcal{S}_{i,j\in kC} = \sum_{k'} x_i ^{\top} w_j^{k'} \frac{exp(x_i ^{\top}w_j^{k'})}{\sum_{k'} exp(x_i ^{\top}w_j^{k'})}
    \label{eq:cassignment}
\end{equation}

Softtriple only generates disjoint separable clusters in the 
latent space, and does not smooth the latent space ($\lambda$ performs label smoothing). 
We add a regularizer $L_{smooth}$ to smooth the intra-proxy space of the embeddings 
to achieve a more convex hull while maintaining accuracy at the cluster boundaries. 
Existing regularizers such as 1-Lipschitzness or KL-divergence enforce smoothness 
on the entire latent space without considering high-density disjoint clusters. 
We propose a contrastive smoothing regularizer that explicitly works with 
the softtriple clusters to improve smoothing in the high-density 
regions while ignoring low-density regions. 
We do this by minimizing the entropy between intra-proxy samples only, with:

\begin{equation}
    \mathscr{L}_{smooth} = \frac{1}{n} \sum_{x,y\sim\mathscr{D}} \max_{x_i, x_j|y_i^{k'}=y_j^{k'}}  D_{SKL}(f(x_i), f(x_j)) 
    \label{eq:lsmooth}
\end{equation}

Here $f(x_i)$ yields a probability simplex for the embeddings, 
and $D_{SKL}$ is the symmetrized KL divergence between $x_i$ and $x_j$ 
if both belong to the same proxy cluster. 
Compared to SMART \cite{smart}, which minimizes a Lipschitz smoothness 
across the entire latent space, our approach focuses specifically on high-density regions. 
Our complete loss, with tuning parameters and combining with \autoref{eq:lft} is:

\begin{equation}
    \mathscr{L} = \mathscr{L}_{FT} + \lambda_{smooth}\mathscr{L}_{smooth}
\label{eq:lfinal}
\end{equation}

\begin{table}[t]
    \scriptsize
    \begin{tabular}{lrrrr|rr}
    \hline
    \multicolumn{1}{r}{}                          & \multicolumn{4}{c|}{AlBERT}                                                                                                 & \multicolumn{2}{c}{p-value (vs UFIT)} \\
    \multicolumn{1}{r}{\multirow{-2}{*}{Dataset}} & PTM       & Masker                           & SMART                            & UFIT                                      & Masker             & SMART            \\ \hline
    rumor                                         & 0.55±0.04 & 0.63±0.05                        & 0.65±0.08                        & \textbf{0.76±0.05}                        & 0.003              & 0.031            \\
    miscov                                        & -         & {\red{ 0.49±0.01}} & {\red{0.48±0.02}} & {\red{\textbf{0.56±0.04}}} & 0.005              & 0.004            \\
    covid\_fn                                     & -         & {\red{0.61±0.08}} & 0.65±0.06                        & \textbf{0.86±0.03}                        & 0.000              & 0.000            \\
    coaid                                         & -         & 0.61±0.03                        & 0.61±0.04                        & \textbf{0.74±0.04}                        & 0.000              & 0.001            \\
    covid\_cq                                     & -         & {\red{0.53±0.02}} & {\red{0.52±0.06}} & \textbf{0.63±0.05}                        & 0.003              & 0.014            \\ \hline
    \end{tabular}
    \caption{\textbf{Never-seen-before Novelty}. PTM$\rightarrow$\ttt{efnd}: an AlBERT model without finetuning is tested on all \ttt{efnd} datasets 
    for baseline OOD accuracy. 
    Each ID column indicates a fine-tuned classifier trained 
    and tested on the Dataset. 
    Each ND column is a classifier fine-tuned on Dataset and 
    tested on \{\ttt{efnd}\}\textbackslash{}Dataset. \red{Red} values are either worse than PTM$\rightarrow$\ttt{efnd} or do not pass t-test.}
    \label{tab:efnd}
\end{table}

\PP{Training Details} We implemented our code on PyTorch with the EdnaML framework \cite{ednaml}; 
our code is available at [anonymized-link]. 
During training, we used batch sizes of 64; learning rate of 1e-4 
with a warmup for the first half-epoch from 1e-5 and 
decay of 0.6 every 2 epochs; most classifiers trained for at 
most 10 epochs with early stopping. 
For losses, we used $\lambda_{smooth}=1$, and $\lambda_{MLM}=\lambda_{KL}=0.5$.

%% file: results.tex
\section{Experimental Evaluation}
\label{sec:results}

\subsection{Limited Incremental Evaluation Approach}
\label{sec:datasets}
Misinformation detection research uses primarily 
fixed datasets\cite{SurveyFNZhou2020,SurveyFNDataSets2021}, 
\eg using the gold standard k-fold cross-validation in carrying out ID or OOD tests for 
generalizability\cite{generalizability}. 
In contrast, the continuously arriving new-normal data sets 
separates the training data (from previous campaigns) from test data 
(never-seen-before novelty), similar to domain generalization\cite{DGSurvey2022}. 
Our base experiments (see \autoref{sec:motivation}, \autoref{fig:brittle}) start from two domains that have never-seen-before 
novelty w.r.t. each other: \texttt{coaid} and \ttt{rumor} from EFND \cite{generalizability}. 
These are consistent with typical domain generalization papers with leave-one-domain-out experiments on two domains. 

For subsequent experiments, we deviate from the leave-one-out approach, 
which assumes all domains are fixed and available. 
Following the real-world (real-time arrival) scenario, we evaluate 
classifier $k$ (trained on first $k$ domains) using the $k+1$ test data, 
which classifier $k$ has not seen. 
We call this evaluation method limited incremental evaluation. 
A skeptical reviewer might observe that the base case $(k=1)$ is 
consistent with the leave-one-out approach with two domains.

\begin{figure}[t]
    \centering
    \includegraphics[width=0.95\columnwidth]{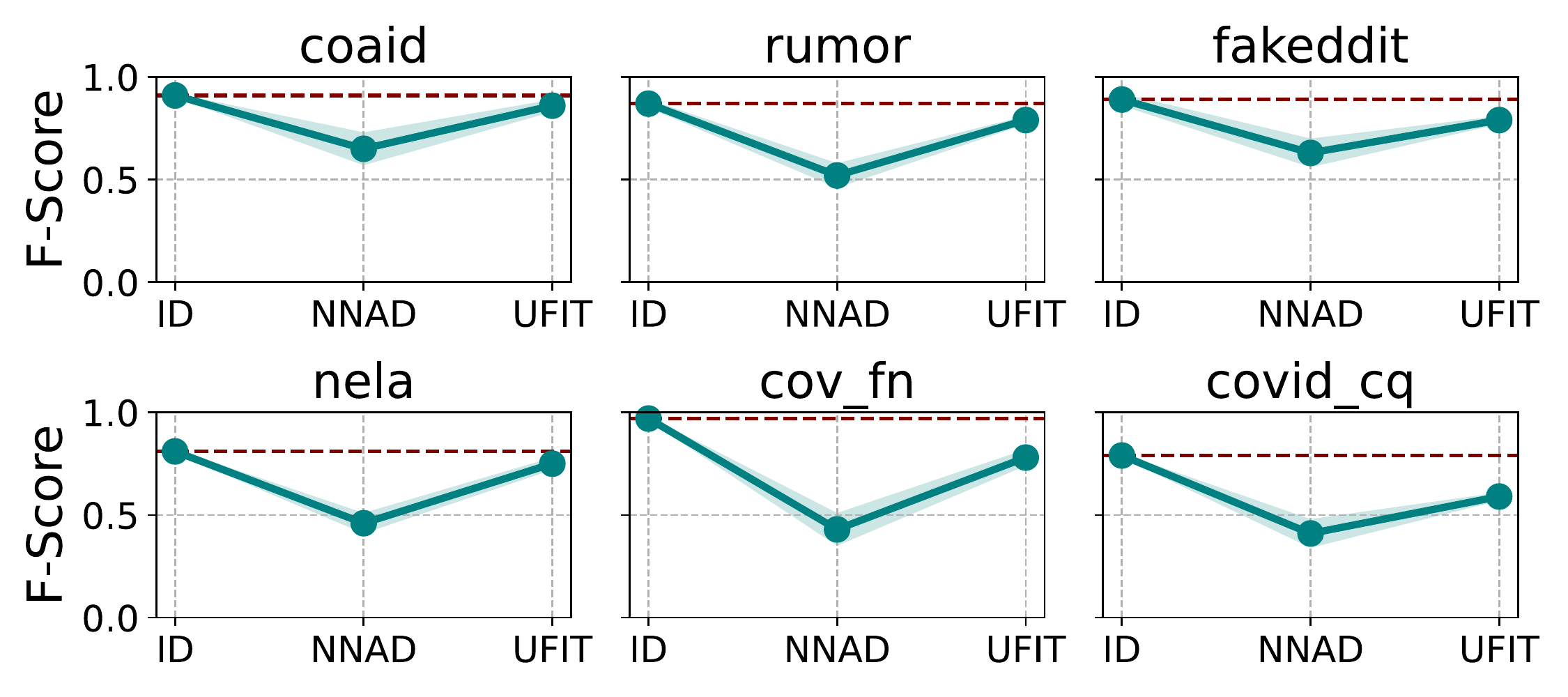} 
    \caption{Impact of \sys across multiple datasets of \ttt{efnd} in recovering ID accuracy in an NNAD setting}
    \label{fig:efnd}
\end{figure}

The difference between incremental evaluation and leave-one-out 
arises and becomes amplified as the number of domains $k$ (new fake news campaigns) grows larger. 
In limited incremental evaluation, classifier $k$ does not 
know about the $k+1$ campaign, which is the real world situation. 
In leave-one-domain-out evaluation, the evaluation results are improved by a 
combination of $k$ experiments, with only the classifier $k$ unaware of the $k+1$ 
data, since the other $k-1$ classifiers all include the $k+1$ knowledge. 
While a common practice, we believe that the import of future 
knowledge ($k+1$ data) reduces the `bigger domain shift' \cite{LiDeeper2017} 
and causes an artificial inflation of combined classifier performance 
in such generalizability experiments.  

In the limited incremental evaluation of new-normal arbitrary 
distribution (NNAD) data sets such as NELA \cite{nela}, 
the generalization experiments are conducted on classifiers that only have 
$k$ data set knowledge, but tested on $k+1$ data sets. 
As comparison, we use an oracle classifier trained with full ($k+1$) 
data knowledge that achieves very high accuracy; see \autoref{sec:nela}. 
%

We perform experiments on the following datasets, grouped into their evaluation area:
\squishlist
\item \textbf{Never-seen-before novelty}. To evaluate \sys with respect to 
this property of new-normal, we use the EFND COVID-19 Fake News dataset from \cite{generalizability}. 
EFND contains 11 public fake news datasets that are semantically similar 
yet exhibit NNAD by capturing different types of misinformation on twitter and news articles in 2020. 
%
\item \textbf{Massive short campaigns}. Here, we use the 4 yearly releases of the NELA dataset spanning 2018 through 2021
as a temporally scoped dataset with changing misinformation topics year-to-year. 
Each yearly release contains misinformation posts collected in that year, 
and exhibit NNAD relative to each other due to changes in misinformation campaigns. 
%
\squishend

\begin{figure}[t]
    \centering
    \includegraphics[width=0.95\columnwidth]{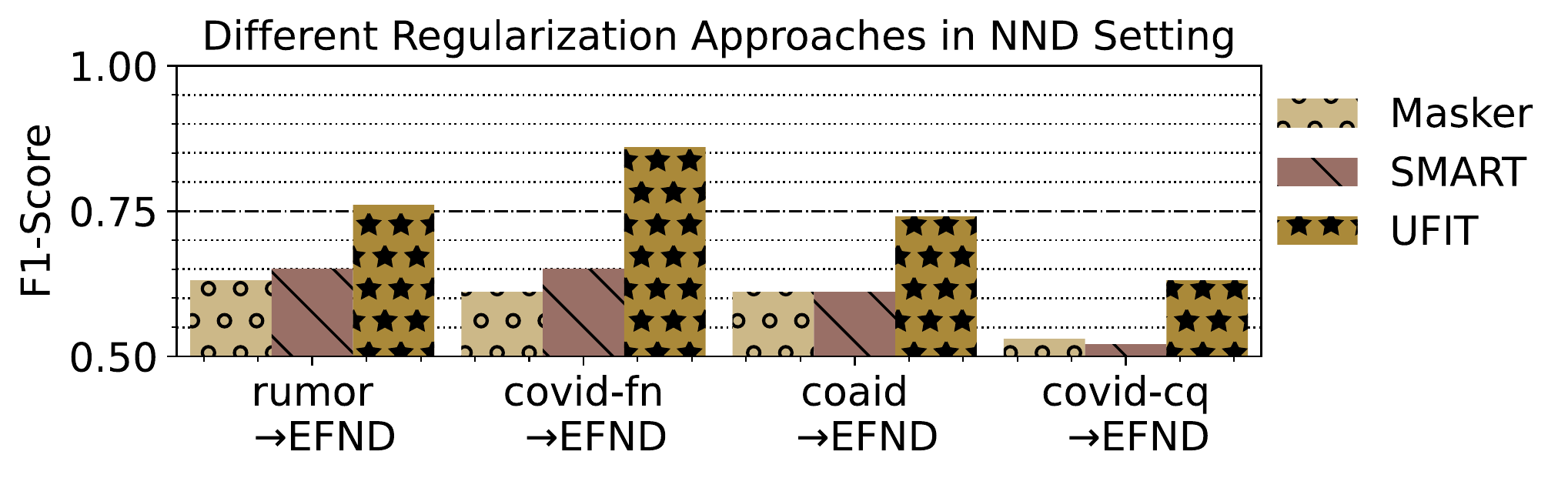} 
    \caption{\sys{'s} approach of combining masking (Masker) and smoothing (SMART), with emphasis on high-density sets, outperforms either individual approach and PTMs in the NNAD setting}
    \label{fig:regdiff}
\end{figure}

\begin{table}[t]
    \centering
    \small
    \begin{tabular}{@{}clrrr@{}}
    \toprule
    \multicolumn{2}{l}{Training subsets}          & NELA-18            & \begin{tabular}[c]{@{}r@{}}NELA-18, \\ NELA-19\end{tabular} & \begin{tabular}[c]{@{}r@{}}NELA-18, \\ NELA-19, \\ NELA-20\end{tabular} \\
    \multicolumn{2}{l}{\textit{Testing   subset}} & \textit{NELA-19}   & \textit{NELA-20}                                            & \textit{NELA-21}                                                        \\ \midrule
    \textbf{Base}        & \textbf{Approach}  & -                  & -                                                           & -                                                                       \\ \midrule
    \multirow{6}{*}{AlBERT}  & PTM                & 0.57$\pm$0.01          & 0.55$\pm$0.04                                                   & 0.54$\pm$0.04                                                               \\
                             & MDAWS              & 0.70$\pm$0.01          & 0.68$\pm$0.04                                                   & 0.63$\pm$0.03                                                               \\
                             & FreeLB             & 0.83$\pm$0.02          & 0.71$\pm$0.02                                                   & 0.72$\pm$0.02                                                               \\
                             & Masker             & {\ul 0.85$\pm$0.03}    & {\ul 0.78$\pm$0.04}                                             & {\ul 0.79$\pm$0.03}                                                         \\
                             & SMART              & 0.84$\pm$0.01          & 0.76$\pm$0.04                                                   & 0.76$\pm$0.01                                                               \\
                             & \textbf{\sys}      & \textbf{0.87$\pm$0.03} & \textbf{0.85$\pm$0.01}                                          & \textbf{0.88$\pm$0.01}                                                      \\ \midrule
    \multirow{6}{*}{BERT}    & PTM                & 0.59$\pm$0.02          & 0.56$\pm$0.03                                                   & 0.56$\pm$0.03                                                               \\
                             & MDAWS              & 0.70$\pm$0.03          & 0.68$\pm$0.04                                                   & 0.65$\pm$0.03                                                               \\
                             & FreeLB             & 0.85$\pm$0.02          & 0.73$\pm$0.03                                                   & 0.72$\pm$0.01                                                               \\
                             & Masker             & 0.86$\pm$0.01          & {\ul 0.80$\pm$0.02}                                             & {\ul 0.81$\pm$0.03}                                                         \\
                             & SMART              & {\ul 0.86$\pm$0.01}    & 0.76$\pm$0.02                                                   & 0.76$\pm$0.02                                                               \\
                             & \textbf{\sys}      & \textbf{0.89$\pm$0.02} & \textbf{0.86$\pm$0.02}                                          & \textbf{0.90$\pm$0.04}                                                      \\ \bottomrule
    \end{tabular}
    
    \caption{Here, we test yearly subsets of NELA across multiple approaches, described in \autoref{sec:nela}. \sys tracks the oracle accuracy with the least decline in performance.}
    \label{tab:nela}
    \end{table}

\begin{figure}[t]
    \centering
    \includegraphics[width=0.95\columnwidth]{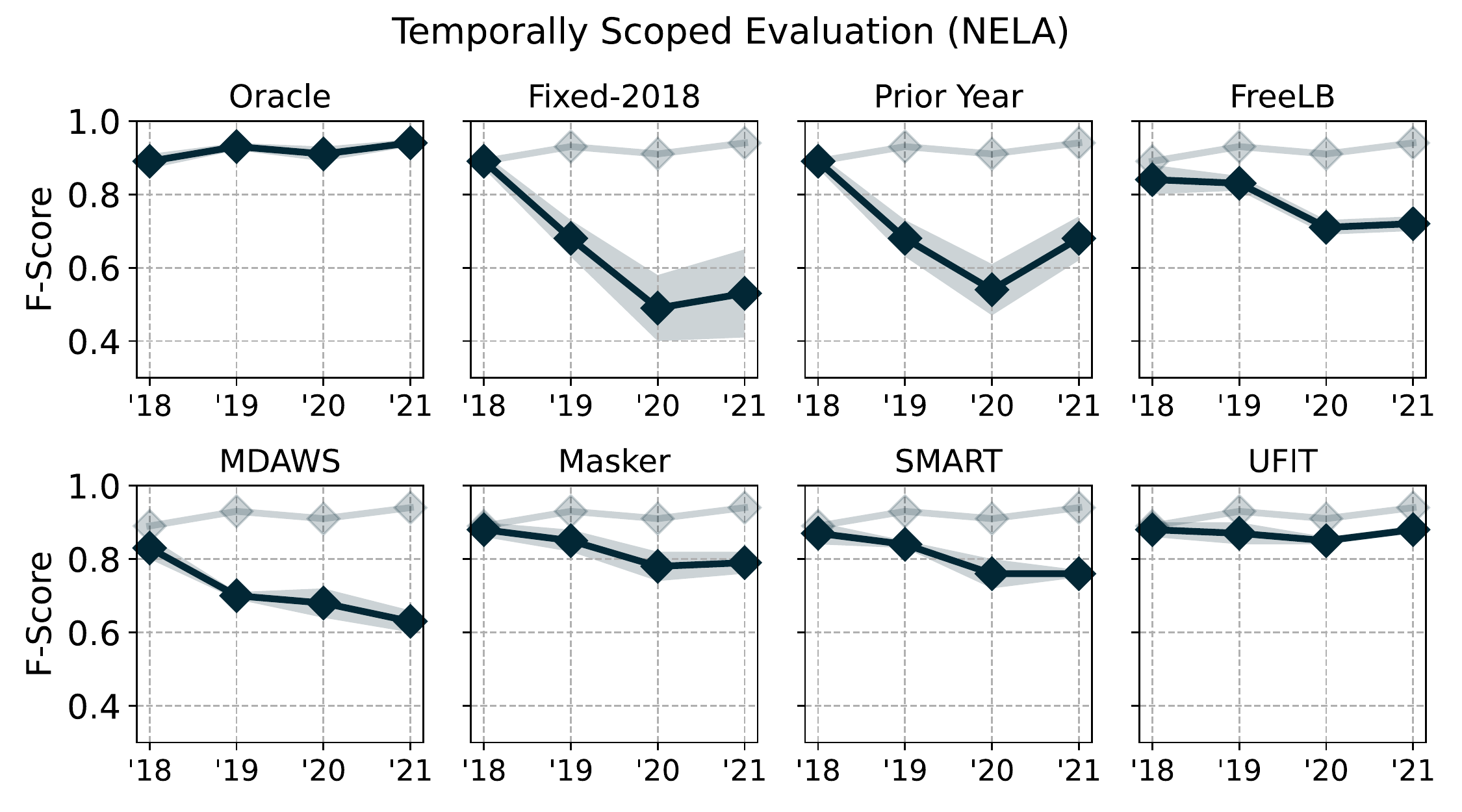} 
    \caption{We compare \sys to several approaches described in \autoref{sec:nela} to an oracle with complete knowledge of all $k+1$ datasets}
    \label{fig:temporal}
\end{figure}

\subsection{Never-seen-before Novelty: \ttt{efnd}}
\label{sec:efnd}
Our key motivation for using EFND to evaluate never-seen-before novelty, 
versus using traditional single-dataset evaluations, is the presence of NNAD between the EFND datasets. 
A thorough exploration of the pairwise novelty conducted in \cite{generalizability} shows 
fine-tuned classifiers trained on any single EFND dataset significantly 
underperform on the remaining datasets. 
With \sys, we show in \autoref{tab:efnd} and \autoref{fig:efnd} that when we train on 
one of the EFND datasets, the classifier's accuracy on remaining 
datasets is significantly higher than PTM. 
We compare to Masker \cite{masker}, which implements only masking, and SMART \cite{smart}, 
which implements only smoothing. 
\sys, which implements both semantic masking and smoothing, 
with the added constraint of high-density sets, outperforms Masker and SMART across multiple EFND datasets
as training data, shown in \autoref{fig:regdiff}.

There are several observations from 
\autoref{tab:efnd}, \autoref{fig:efnd}, and \autoref{fig:regdiff}:
(1) Supporting the conclusions of \cite{robust}, the PTM classifier is 
more reliable than either Masker or SMART due to robustness of 
pre-trained LLMs. 
(2) \sys{'s} ID accuracy is overall lower than SMART's 
ID accuracy, primarily due to underfitting and reducing overall test 
accuracy slightly on in-distribution samples. 
This is also reflected in Masker's lower accuracy on ID vs SMART. 
(3) However, the slightly lower accuracy on ID is counteracted by significantly 
better performance in NNAD setting for \sys, where we evaluate the classifier on the 
remaining EFND datasets. 
Masker sees on average a 20\% drop in accuracy ($\pm$7\%). 
SMART sees 25\% drop in accuracy ($\pm$2\%). 
\sys, on the other hand, sees only 8\% drop in accuracy ($\pm$3\%).

\subsection{Massive Short Campaigns: \ttt{nela}}
\label{sec:nela}
Different from EFND, NELA contains yearly dataset releases; 
the timestamp is an important consideration for massive short campaigns, 
since we wish to explore impact of temporally scoped knowledge on classifier overfitting. 
That is, each year new misinformation topics emerge: NELA-20 contains novel COVID-19 misinformation, 
and NELA-19 contains novel election misinformation in the run-up to the 2020 primary 
elections in the US \cite{nela}. 
Different from EFND, where we have explicitly ensured new-normal properties by 
including multiple datasets, NELA's long-term collection approach 
is sufficient for new-normal properties; we can see fixed classifier deterioration (Fixed-2018) in \autoref{fig:temporal}.

To fit the limited incremental evaluation approach, we take each yearly 
subset of NELA starting from 2019 as testing data and use the prior yearly 
subsets as training data across different backbones (\eg $\{0,\cdots, k-1,k\}\rightarrow k+1$). 
In addition to Masker and SMART, we also tested MDAWS \cite{mdaws} which trains 
a domain-invariant regularizer and FreeLB \cite{freelb} which trains a 
adversarial perturbation-based smoothness regularizer (similar in spirit to SMART and MiDAS \cite{midas}). 

We show our results in \autoref{tab:nela} and \autoref{fig:temporal}; the latter 
contains 2 additional evaluations: Fixed-2018, where we take an AlBERT trained on NELA-2018 
and use it for all years ($k\rightarrow \{k+1, k+2,\cdots\}$), and Prior-Year ($k\rightarrow k+1$), 
which uses an AlBERT trained on each prior year. 
Both experience degraded performance compared to using all available 
prior data (as we do in \autoref{tab:nela}). 
MDAWS, which trains a domain-invariant representation, performs relatively 
poorly on the yearly subsets; we attribute this to the domain invariance 
representation that loses significant domain-specific information. 
Similarly, we attribute FreeLB's underperformance to the fact that 
it enforces smoothness on the latent space everywhere. 
%

With \sys, as before, we have combined masking and smoothness regularizing with the added constraint 
of focusing on the high-density proxies in the latent spaces. 
This combines the best impacts of masking (preventing keyword overfitting) and 
smoothness regularization (improving edge cases accuracy) within the latent space 
the classifier has already modeled.

\begin{figure}[t]
    \centering
    \includegraphics[width=0.95\columnwidth]{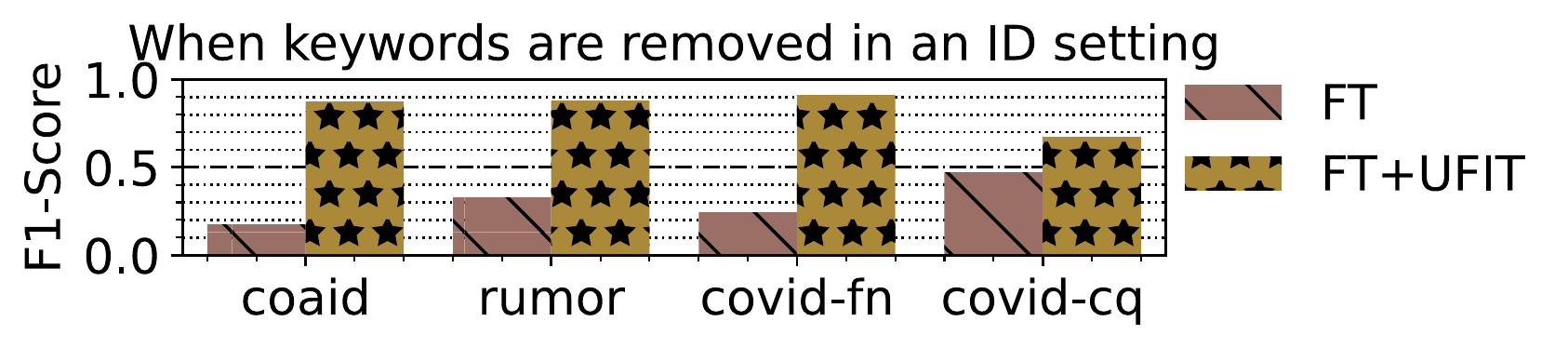} 
    \caption{\sys avoids overfitting on keywords. Conversely, when we mask keywords in an ID setting, FT classifier is worse than random guessing due to adversarial overlap in misinformation.}
    \label{fig:overfitting}
\end{figure}

\subsection{Analysis of Controlled Underfitting}
\label{sec:analysis}

\PP{Reducing Overfitting} We show that \sys explicitly reduces overfitting 
with a straightforward experiment. 
We take a fine-tuned model without UFIT (FT) and with UFIT (FT+UFIT) 
and test them in an ID setting; from the testing data, we mask the keywords identified by UFIT. 
\autoref{fig:overfitting} shows this experiment with EFND, where the x-axis is the training dataset.
FT has significant deterioration when keywords are masked, compared to \sys, which maintains high reliability.

\begin{figure}[t]
    \centering
    \includegraphics[width=0.95\columnwidth]{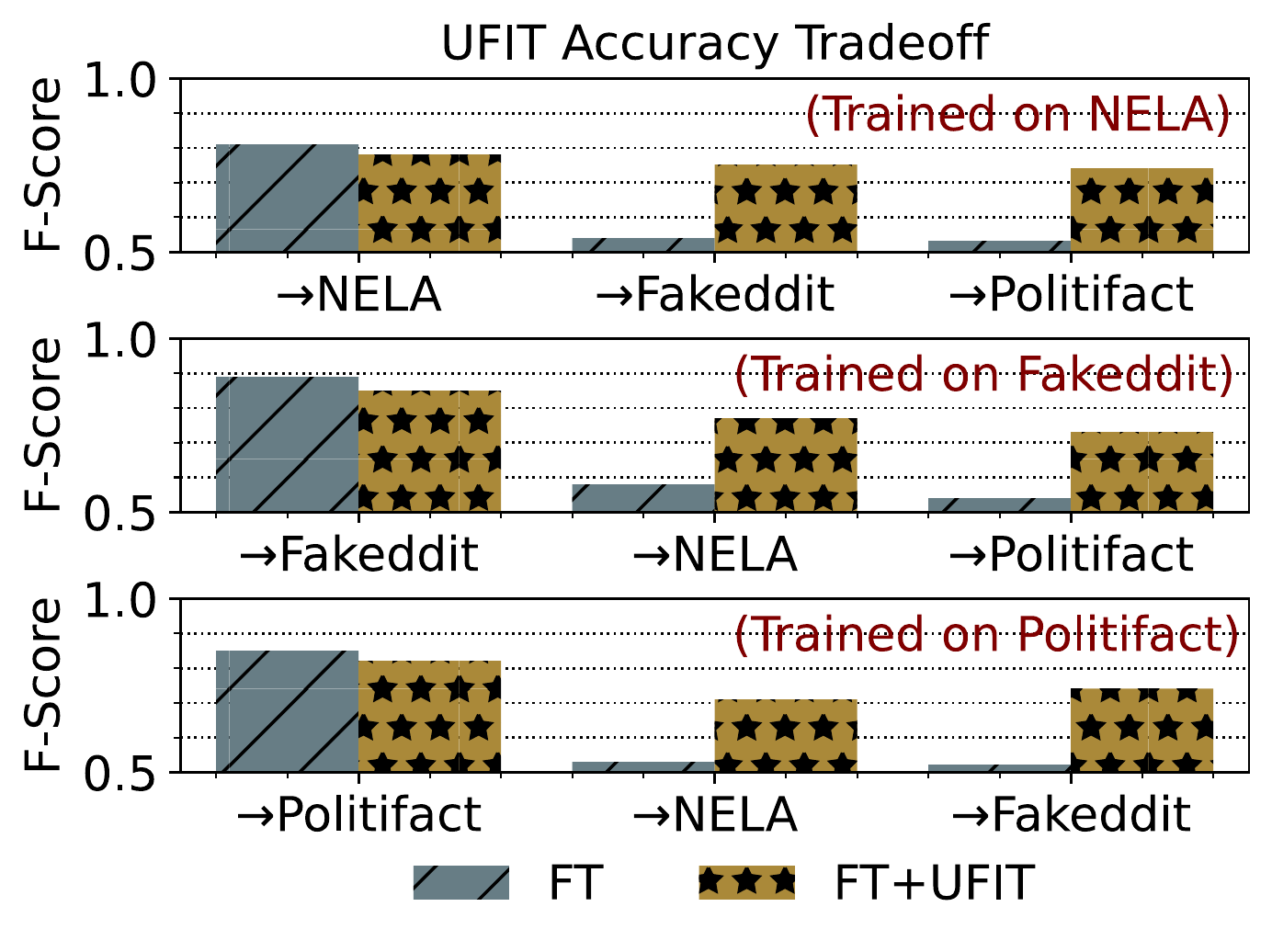} 
    \caption{\sys has significantly more reliable performance in the NNAD setting, with negligibly lower accuracy in ID setting (due to underfitting) }
    \label{fig:tradeoff}
\end{figure}

\PP{Accuracy Tradeoff} \sys has negligibly lower accuracy ($\sim$~1-5\%) 
in an ID setting compared to a fine-tuned classifier that has overfit 
on the training data. 
In the NNAD setting, UFIT significantly outperforms the overfit variants by $\sim$25\%, 
due to combination of masking and regularization that improve generalized 
feature selection for misinformation detection in the latent space. 
We summarize these results in \autoref{fig:tradeoff} with NNAD experiments using NELA, Fakeddit \cite{fakeddit}, and Politifact \cite{fnn}.

\begin{table}[t]
    \centering
    \tiny
    \begin{tabular}{crrrr}
    \hline
    \textbf{Testing   Dataset} & \textbf{\begin{tabular}[c]{@{}r@{}}PTM L-Score\\      PTM Acc.\end{tabular}} & \textbf{\begin{tabular}[c]{@{}r@{}}FT-ID L-Score\\      FT-ID Acc.\end{tabular}} & \textbf{\begin{tabular}[c]{@{}r@{}}FT-NNAD L-Score\\      FT-NNAD Acc.\end{tabular}} & \textbf{\begin{tabular}[c]{@{}r@{}}UFIT L-Score\\      UFIT Acc.\end{tabular}} \\[3mm] \hline
    \ttt{coaid}                      & \begin{tabular}[c]{@{}r@{}}21.4$\pm$1.4\\      0.63$\pm$0.01\end{tabular}                & \textbf{\begin{tabular}[c]{@{}r@{}}6.5$\pm$0.8\\      0.91$\pm$0.01\end{tabular}}      & \begin{tabular}[c]{@{}r@{}}35.4$\pm$5.3\\      0.65$\pm$0.08\end{tabular}              & \begin{tabular}[c]{@{}r@{}}\underline{10.3$\pm$2.8}\\      \underline{0.86$\pm$0.03}\end{tabular}                  \\[3mm]
    \ttt{cov\_rumor}                 & \begin{tabular}[c]{@{}r@{}}26.4$\pm$2.9\\      0.67$\pm$0.02\end{tabular}                & \textbf{\begin{tabular}[c]{@{}r@{}}8.3$\pm$0.7\\      0.87$\pm$0.02\end{tabular}}      & \begin{tabular}[c]{@{}r@{}}41.5$\pm$4.6\\      0.52$\pm$0.06\end{tabular}              & \begin{tabular}[c]{@{}r@{}}\underline{12.6$\pm$3.1}\\      \underline{0.79$\pm$0.02}\end{tabular}                  \\[3mm]
    \ttt{covid\_cq}                  & \begin{tabular}[c]{@{}r@{}}27.9$\pm$1.6\\      0.56$\pm$0.02\end{tabular}                & \textbf{\begin{tabular}[c]{@{}r@{}}7.1$\pm$0.6\\      0.85$\pm$0.02\end{tabular}}      & \begin{tabular}[c]{@{}r@{}}86.5$\pm$9.1\\      0.48$\pm$0.09\end{tabular}              & \begin{tabular}[c]{@{}r@{}}\underline{16.1$\pm$3.5}\\      \underline{0.76$\pm$0.03}\end{tabular}                  \\[3mm] \hline
    Summary                    & \begin{tabular}[c]{@{}r@{}}Baseline L\\      Baseline Acc.\end{tabular}          & \begin{tabular}[c]{@{}r@{}}Very Low L\\      High Acc.\end{tabular}            & \begin{tabular}[c]{@{}r@{}}Very High L\\      Low Acc.\end{tabular}            & \begin{tabular}[c]{@{}r@{}}Low L\\      Good Acc.\end{tabular}                     \\ \hline
    \end{tabular}
    \caption{Lipschitz scores (L-score) and Accuracy of PTM, FT, and FT+UFIT. L-score is computed by taking max of all proxy Lipschitz scores. Low L = More smooth.}
    \label{tab:lipschitz}
\end{table}

\PP{Intra-Proxy Regularization}
We show the direct impact of smoothing by computing local Lipschitz 
scores \cite{niceness} for PTM, 
fine-tuned classifier (FT) in an ID setting and NNAD setting, and FT+UFIT in an NNAD setting, plus corresponding accuracy
in \autoref{tab:lipschitz}.
FT-ID has the lowest 
L-score, indicating smoothest latent space around the testing data. 
FT in NNAD setting, due to overfitting on training data, 
has the highest L-score (relative to PTM as well), and lowest accuracy, 
indicating poor generalization on testing data. 
\sys achieves higher smoothness (lower L score) than FT in NNAD due to regularization 
so that even with new-normal, it maintains high accuracy relative to FT in ID setting.

%% file: conclusions.tex
\section{Conclusion}
\label{sec:conc}
Misinformation exhibits two key properties we call new-normal: 
(i) never-seen-before novelty, which creates transient uncertainty in classifier predictions 
by introducing novel text vectors for fake news, and 
(ii) massive short campaigns, where new misinformation continuously 
replaces existing topics in the stream. 
Existing misinformation detection approaches that use a single 
gold dataset for training and testing are limited under 
these new-normal challenges, since they can overfit to non-generalizable 
strong signals in their training data. 
We proposed \sys, a controlled underfitting method to carefully reduce 
the degree of overfitting in classifiers to adapt them for the new-normal setting of misinformation detection.

\sys combines semantic masking and smoothness regularization in distinct 
high-density clusters in the latent space and improves reliability 
in the new-normal setting by creating classifiers that do not have explicit 
over-reliance on strong signals in training data. 
\sys differs from most regularization methods by 
explicitly underfitting high-density regions of the latent space 
where classifiers tend to overfit. 
%

\sys is a promising solution for addressing overfitting in fine-tuned 
classifiers in the NNAD setting, particularly in adversarial domains such as misinformation detection where 
reliability is crucial. 
We hope \sys will aid in further research on quantifying and detecting new-normal 
properties as and when they occur, as well as developing efficient techniques for such scenarios.